\pdfoutput=1

\documentclass[11pt]{article}

\usepackage[]{acl}

\usepackage{times}
\usepackage{latexsym}

\usepackage[T1]{fontenc}

\usepackage[utf8]{inputenc}

\usepackage{microtype}

\usepackage{graphicx}
\usepackage{amsmath}
\usepackage{booktabs} 
\usepackage[table]{colortbl}
\colorlet{mycell}{blue!25}
\usepackage{pifont}
\usepackage{fancyhdr}

\newcommand{\myVmark}{\ding{51}}
\newcommand{\myXmark}{\ding{53}}

\usepackage{enumitem,amssymb}
\newlist{todolist}{itemize}{2}
\setlist[todolist]{label=$\square$}

\title{Generative Spoken Dialogue Language Modeling}

\author{Tu Anh Nguyen$^{1,3}$, Eugene Kharitonov$^1$, Jade Copet$^1$,  Yossi Adi$^1$, \\ \textbf{Wei-Ning Hsu$^1$, Ali Elkahky$^1$, Paden Tomasello$^1$, Robin Algayres${^1}$, } \\ \textbf{Benoît Sagot$^{3}$, Abdelrahman Mohamed$^{1}$, Emmanuel Dupoux$^{1,2}$} \\
        Meta AI Research$^1$,  
        EHESS, Paris$^2$, Inria, Paris$^3$\\
        \texttt{\{ntuanh, abdo, dpx\}@fb.com}
        }
\usepackage{verbatim}

\begin{document}

\maketitle
\begin{abstract}
We introduce dGSLM, the first ``textless'' model able to generate audio samples of naturalistic spoken dialogues.  It uses recent work on unsupervised spoken unit discovery coupled with a dual-tower transformer architecture with cross-attention trained on 2000 hours of two-channel raw conversational audio (Fisher dataset) without any text or labels. 
We show that our model is able to
generate speech, laughter and other paralinguistic signals in the two channels simultaneously and reproduces more naturalistic and fluid turn taking compared to a text-based cascaded model\footnote{\label{demowebsite}Generation samples can be found at \url{https://speechbot.github.io/dgslm}}\footnote{Code and pre-trained models will be made available at \url{https://github.com/facebookresearch/fairseq/tree/main/examples/textless_nlp/dgslm}}.
\end{abstract}

\section{Introduction}
In natural conversations, speakers spontaneously coordinate who is currently speaking and when the other person will speak next. As a result, conversations end up being a fluent succession of \textit{turns} without much overlapping speech or long stretches of silence. Of course, silences and overlaps also occur naturally and they carry significant information which is interpreted within the conversation setting. For instance, when overlapping speech occurs it often contains content-neutral verbal information (e.g., "hmm", "yeah") or non-verbal vocalization (e.g., laughter), used to convey a listening attitude (back-chanelling) \cite{yngve1970getting,schegloff1982discourse}. Short silences between turns do occur and show both cross-cultural variations and universal dependence on dialogue related variables, for instance, straight and positive answers to questions are typically faster than non-responses or negative responses \cite{stivers2009universals}.

All of this \textit{turn-taking} coordination is natural to humans, and starts to be learned at an early age by infants \cite{nguyensystematic}. In contrast, it remains a challenging area of research in human/machine interactions \cite{skantze2021turn}. One of the reason is that much of the research into natural dialogue modeling is taking place with text-based interfaces. Here, the coordination problem is primarily focused on semantic coherence and appropriateness of the artificial agent in interaction with a human (see \citealp{ni2021dialoguereview} for a review). The turn-taking problem itself is being taken care of by an artificially imposed walkie talkie arrangement; each agent is writing in turn and signalling the end of it's turn by pressing carriage return. 

Within speech-based systems, it is very similar, as current spoken assistants like Siri or Alexa are triggered by a predetermined wake word, and wait for the end of an utterance followed by sufficient silence to segment the turns of the human interlocutor. This may give rise to slow and unnatural conversations. In fact, in human-human conversation, \textit{pauses} within speaker turns tend to be on average longer than \textit{gaps} between speaker turns \cite{brady1968statistical,ten2005temporal, heldner2010pauses}, indicating that silence may not be the main cue for humans to switch turns.
Because most speech-based systems are based on Automatic Speech Recognition (ASR), and that many significant aspects of speech like prosody and nonverbals are typically not annotated in naturalistic speech dialogues, current dialogue systems have been struggling with generating naturalistic dialogue.

Here we capitalize on recent progress in self-supervised learning and textless speech processing \cite{borgholt2022brief,borsos22_audiolm,kushal2021gslm} to investigate the possibility to directly train a spoken dialogue model from raw audio, bypassing the need for text or ASR. 
Briefly, we build on self-supervised discrete speech representations models, which we train on spontaneous conversations with each speaker having his or her own audio channel.
After training, the speech units come to represent not only verbal but also nonverbal materials. We can now encode a conversation between two interlocutors as two parallel streams of discrete tokens. We then introduce a novel dual-tower transformer architecture, where each channel is processed by one "tower" of the model which learn via an autoregressive loss, but the two towers also communicate via cross-attention in their hidden units. This cross-attention is critical for the correct synchronization of the two channels and result in a naturalistic distribution of turns, overlap and pauses. While this system is not trained on enough data to capture deep syntactic and semantic aspects of dialogue, and indeed scores below a text-based cascaded ASR+LM+TTS model on semantic content, it does capture better surface characteristics of chitchat in mimicking accurately turn-taking and backchanneling. This can be seen as a proof of principle that previously difficult to capture aspects of spontaneous conversations can be captured with minimally modified language modeling techniques. Finally, our model opens up new possibilities to create more natural naturalistic human-machine dialogue systems in the future.

\begin{figure}[t]
     \centering
     \includegraphics[width=.3\textwidth]{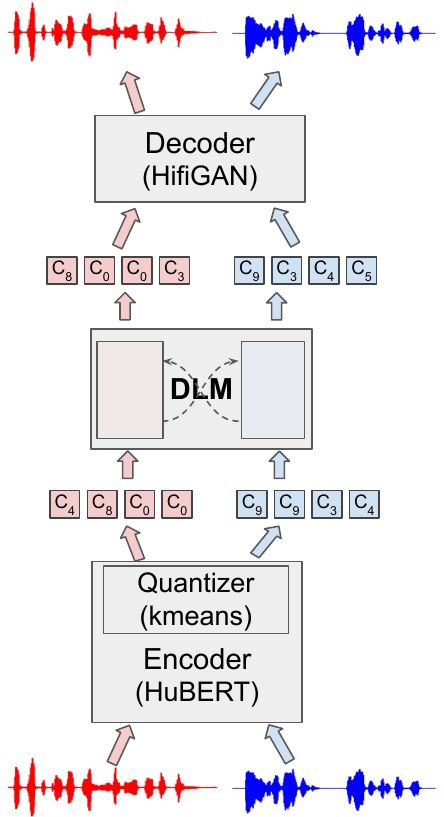}
     \caption{General Schema for dGSLM: A discrete encoder (HuBERT+kmeans) turns each channel of a dialogue into a string of discrete units $(c_1, .. c_N)$. A Dialogue Language Model (DLM) is trained to autoregressively produce units that are turned into waveforms using a decoder (HifiGAN).}
     \label{fig:setting}
\end{figure}


\section{Related work}\label{sec:related}
\paragraph{Unsupervised Spoken Language Modeling.}
Recently great advances have been achieved in the area of representation learning from raw audio. Models trained with either autoencoder objectives \cite{OndelBC16,vqvae} or masked objectives (CPC:  \citealp{oord2018cpc}; APC: \citealp{chung2020generative}; wav2vec 2.0: \citealp{baevski2020wav2vec}; HuBERT: \citealp{weining2021hubert}; MockingJay: \citealp{liu2020mockingjay}) from raw speech can learn audio representation that can be used for a variety of downstream tasks \cite{yang2021superb}, see \citet{borgholt2022brief} for a review.

Most of these models build a codebook of discrete units, either as latent representation or as targets. The discrete representation can in turn be fed to a standard autoregressive language model, which can then be sampled to generate new speech sequences \cite{kushal2021gslm,dieleman2021variable}. An interesting aspect of this procedure is that it can capture aspects of speech that are typically not available in written transcriptions and can therefore model prosody and intonation \cite{Kharitonov2021pgslm}, or non verbal vocalizations typical of emotional speech \cite{kreuk2021textless}. Up to now, however, no such model has been applied to multi-party conversational speech.


\paragraph{Dialogue Generation.}
Since the early work on end-to-end neural dialogue generation~\cite{vinyals_dialogue_15, Div_dialogue, dialogue_review}, empowered by scalable methods for language representation~\cite{gpt, BART}, there has been enormous progress in the area of dialogue generation~\cite{BlenderBot, DialoGPT, Meena}. More recent research focused on utilizing retrieval augmented generation methods~\cite{rag} for long-context, multi-session conversations~\cite{longdialogue}, and grounding responses on fresh information from the internet~\cite{internetdialogue,seeker}. However, all the progress in this research work centered around text dialogues leaving out non-lexical information~\cite{SCHULLER20134, Ang2002ProsodybasedAD} in human-human dialogues, e.g., emotion, pauses, laughter, hesitation, and interruption. Our work builds on end-to-end techniques while taking a speech-first approach to address this shortcoming, where prompts and generated sequences are represented as self-supervised discrete speech representations~\cite{kushal2021gslm}. As a result, the capacity of our models is constrained by the amount of publicly available speech dialogues; for example, the LDC English Fisher dialogues corpus~\cite{Cieri2004TheFC} contains roughly 12M words compared to tens of billions of words in the case of text-based dialogue systems. There have been recent calls for large-scale end-to-end benchmarks and datasets with spoken input to fill this gap~\cite{speechNLU}.

\paragraph{Turn-taking Modeling.}

Decades-long research on conversation analysis 
\cite{duncan1972some, sacks1974simplest, schegloff2000overlapping, gravano2011turn, levinson2015timing, ward2019prosodic}
has shown that human turn-taking relies on a variety of complex signals, or cues, including prosodic cues, linguistic cues and even non-verbal cues such as gaze or gestures, making turn-taking modeling a challenging problem. Simple turn-taking models  using finite-state machines have been proposed to predict the distribution and durations of turn-taking events 
\cite{cassell2001human, Thorisson02naturalturn-taking, raux2009finite}.
More recently, more sophisticated machine learning-based models of turn-taking have been introduced 
\cite{meena2014data, skantze2017towards, roddy2018investigating, masumura2018neural}.
These models used multi-modal features including simple linguistic features and prosodic features extracted from the speech to predict turn shifts. Most recently, \citet{ekstedt2020turngpt} has shown the possibility of turn-taking prediction in spoken dialogue using only linguistics features (text input). We use these definitions of turn-taking events to analyse the output of our models. 

\begin{figure*}[t]
     \centering
     \includegraphics[width=1\textwidth]{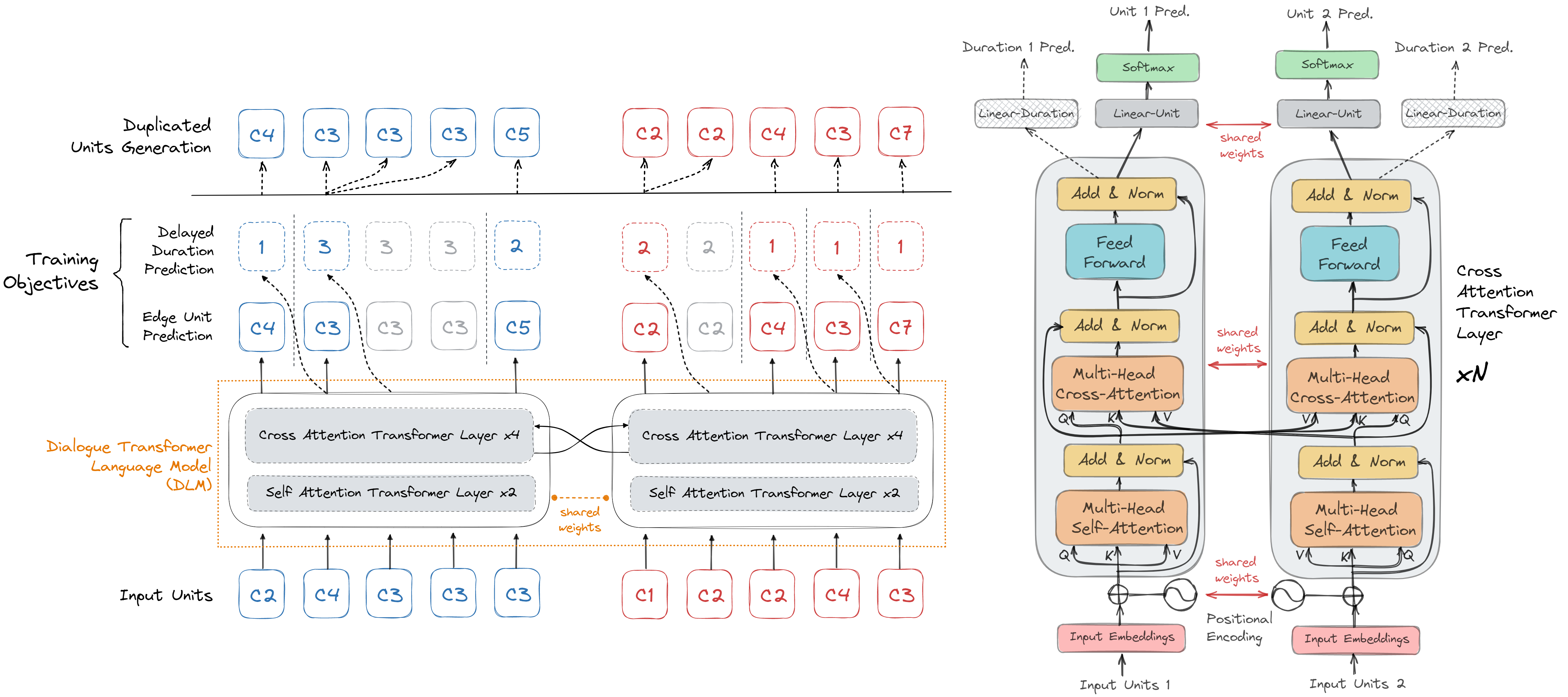}
     \caption{\textbf{Illustration of the Dialogue Transformer Language Model (DLM).} Left:  DLM Training Objectives. During training, the loss is applied only to edge units and their durations. During generation, the model duplicates the units with the corresponding predicted durations. Right: The Cross-Attention Transformer Layer Architecture.}
     \label{fig:arch}
\end{figure*}

\section{Approach}

Our approach is based on the availability of a dataset constructed along the Fisher Telephone conversation collection protocol \cite{Cieri2004TheFC} where each conversation involves two speakers, and each speaker is recorded in a separate audio channel while having a very casual conversation. We follow the textless generative spoken language modeling pipeline of \citet{kushal2021gslm}, which decomposes the problem of speech generation into three components: a Speech-to-Units encoder, a Units-to-Units language model and a Units-to-Speech decoder. For the encoder we adopt HuBERT, \cite{weining2021hubert} followed by k-means clustering; 
for the decoder network we use a modified Hifi-GAN neural vocoder \cite{kong2020hifi}, similarly to~\citet{polyak2021speechresynthesis}. 
These models are trained on single channel data from the Fisher dataset and applied to each channel separately, which do not model cross-channel interactions. For the language model, we introduce our new Dialogue Transformer Language Model, or DLM. Figure \ref{fig:setting} presents an overview of our system. The following sections (Sections \ref{sec:app_encoder}--\ref{sec:app_lm}) will present at a high level each component of our model and review the turn-taking terminology in this study (Section \ref{sec:turn-taking-definition}).

\subsection{Discrete Phonetic Representation}\label{sec:app_encoder}
Conversational speech contains casual expressions (filler words like 'hmm') and a variety of non verbal sounds (e.g., laughter) that do not appear in formal or read speech. We therefore train a HuBERT model \cite{weining2021hubert} directly on our conversation dataset in order to obtain domain-appropriate phonetic representation (see Appendix Section \ref{sec:appendix_hubert} for an analysis). Specifically, it is trained on the collection of voice segments extracted of all speakers in the dataset. 
The discrete units are then obtained by clustering the representation of the HuBERT model using the k-means algorithm.
At inference time, the two-channel speech waveform is encoded channel-wise into two time-aligned streams of discrete units.


\subsection{Waveform Generation}\label{sec:app_decoder}
For the waveform generation, we used the discrete unit-based HiFi-GAN vocoder from \citet{polyak2021speechresynthesis} trained on a small subset of high quality single-channel voice segments of our conversation dataset, using discrete units obtained from the HuBERT model and 1-hot speaker information from the dataset. 
During generation, we generate each channel of discrete units with one different speaker, and combine the audio generated from the two channels. Voices for the waveform generation are chosen from the speakers in the HifiGAN training set.

\subsection{Dialogue Transformer Language Model}\label{sec:app_lm}




We introduce our Dialogue Transformer Language Model (DLM), which is
a two-tower transformer network with \textit{Cross-Attention} and shared weights trained with \textit{Edge Unit Prediction} and \textit{Delayed Duration Prediction} objectives. The model is illustrated in Figure \ref{fig:arch} and its components will be detailed below, and we will perform ablations to test for the effects of each of these components. 

We will also compare the two-tower model with a simpler single-tower model with dual inputs. This last model is inspired by previous work in multi-stream language model \cite{Kharitonov2021pgslm}. It consists of a single transformer, with two embedding heads in the input and two softmax heads in the output. This model combines very early the two speaker channels at the embedding layer and models them jointly, only to separate them again in the last layer. We call this model MS-TLM (Multi-Stream Transformer Language Model)

\paragraph{Cross-Attention Transformer Layer.}

When modeling separate channels of dialogue, we would like the LM to not only get information from the history of each channel itself, but also have information from other channels as well. As a result, we add an additional Muti-Head Cross-Attention block after the Multi-Head Self-Attention block to share information between different channels (cf. Figure \ref{fig:arch}, right). 
We train a single Transformer model which we clone into the two towers with shared weights, which allows the model to be speaker-independent without having to do permutation invariant training.

\paragraph{Edge Unit Prediction.}
Previous work \cite{Kharitonov2021pgslm} disentangles the content modeling problem from the duration modeling problem by
training the language model on deduplicated discrete units and the corresponding unit durations with different objectives. 
However, in our setting, units from different channels are time-aligned and there would be no easy way to keep the alignment if we were to deduplicate each input stream. 
On the other hand, 
training a language model on duplicated units is more difficult as
content and duration information are entangled and learnt simultaneously, resulting in a poor modeling performance.
From this point of view, we introduce an edge unit prediction objective, which forces the model to predict the next unit only if it is different from the current one (i.e. edge unit).
We use cross-entropy loss for this objective, and the edge unit prediction loss is then defined as:
\begin{equation*}
    \mathcal{L}_{EU} = \sum_{c=1}^{2}\sum_{\substack{t\\u_t^{(c)} \neq u_{t-1}^{(c)}}}\log p(u^{(c)}_t \mid u^{(1,2)}_{1:t-1}; \theta),
\end{equation*}
where $u_t^{(c)}$ represents the discrete unit from channel $c$ at time $t$ and $\theta$ denotes the model parameters.

\begin{figure*}[ht]
     \centering
     \includegraphics[width=.85\textwidth]{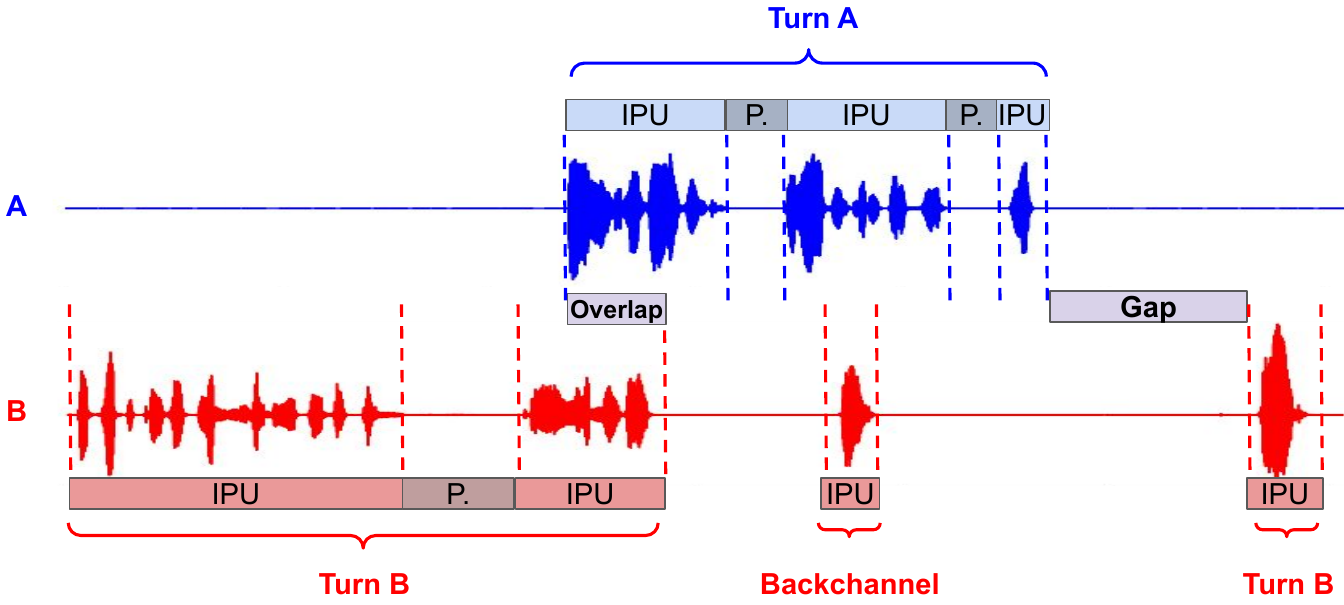}
     \caption{\textbf{Illustration  of turn-taking events:} IPU (Interpausal Unit), Turn (for speaker A and Speaker B, resp), P. (within-speaker Pause), Gap, Overlap and Backchannel. }
     \label{fig:turntaking}
\end{figure*}

\paragraph{Delayed Duration Prediction.}
Besides the unit prediction objective, DLM
models the duration of the edge units
with a duration prediction objective.
As unit durations are highly varied,
we output a continuous duration prediction and employ an L1 loss.
Due to the high correlation between the duration and the unit itself, we follow \citet{Kharitonov2021pgslm} and perform a delayed unit duration prediction, which predicts the duration of an edge unit at time $t$ given the first $t-1+\Delta$ units, where $\Delta$ is a delay factor ($\Delta \geq 0$). 
The delayed duration prediction loss is then defined as:
\begin{equation*}
    \mathcal{L}_{ED} = \sum_{c=1}^{2}\sum_{\substack{t\\u_t^{(c)} \neq u_{t-1}^{(c)}}} \left\lvert  d^{(c)}_t - \hat{d}^{(c)}_t\left(u^{(1,2)}_{1:t-1+\Delta}; \theta\right) \right\rvert,
\end{equation*}
where $d_t^{(c)}$ represents the target duration (number of repetitions) of the edge unit $u_t^{(c)}$ and $\hat{d}^{(c)}_t$ is the continuous duration prediction of the DLM model. 

\paragraph{Training objective.}
The training loss of DLM is the sum of the edge unit prediction loss and the delayed duration prediction loss:
\begin{equation}
    \mathcal{L}_{DLM} = \mathcal{L}_{EU} + \mathcal{L}_{ED}.
\end{equation}

\paragraph{Model Inference for Generation.}
For generation, we autoregressively generate 
edge units and the corresponding durations in both channels. Even though the loss is applied only at the edge units, the model may generate spurious and inconsistent data at other non-edge time steps. We give precedence to the predicted duration associated with the first edge unit predicted in each channel and overwrite the network output with this edge units for the corresponding number of steps. It is this overwritten content which is used as input to the network till the next edge unit. For example, if we predict a unit $u_t^{(c)}$ at time $t$ and the corresponding duration $d_t^{(c)}$ at time $t+1$, we replace the next $d_t^{(c)}$ units of channel $c$ by $u_t^{(c)}$ and only alter the unit at time $t+d_t^{(c)}$. The duration prediction is rounded during generation.

\subsection{Definitions of turn-taking metrics} \label{sec:turn-taking-definition}
Because our model generates two audio channels in parallel, it is possible to use simple Voice Activity Detection (VAD) tools on the output to derive turn-taking metrics. Following Figure \ref{fig:turntaking}, we define an \textit{Inter-Pausal Unit} (IPU) as continuous stretch of speech in one speaker's channel, delimited by a VAD silence of more than 200ms on both side. We define \textit{silence} as sections of the recording with no voice signals on either channel and \textit{overlap} as sections where there are voice signals on both channels. Silences can be subdivided into \textit{gaps} (when it occurs between two IPUs by distinct speakers) and \textit{pauses} (when they occur for the same speaker). Successive IPUs by the same speaker separated by a pause are regrouped into a \textit{turn}. Overlap could also theoretically be subdivided into \textit{backchannel} (when it is rather short IPU contained within an IPU of the other speaker) and \textit{interruption} (when it starts within an IPU of the other channel and continues after its end), but the exact definition is dependant on high-level linguistic features, which we will not attempt to extract here. In our analysis, we will therefore tally the distribution of duration of IPUs, gaps, pauses and overlaps in the training corpus and in generated dialogues of our various models.

\subsection{Cascaded Dialogue Baseline System}
We compare our textless-based dialogue models with a traditional cascaded dialogue system which consists of an ASR model, followed by a text-based language model and a Text-To-Speech (TTS) module. We first transcribe each channel of the dialogue with the ASR model, we then combine the transcribed text into a turn-based conversation\footnote{\label{textturn}example: \texttt{<A> hi <B> hi how you doing <A> great <B> good good my name is marine}}, we ignore any turns that are completely contained inside an other turn. We train a Transformer Language Model on these conversations and we finally employ a TTS module to synthesize the generated text into a turn-based conversation.

\section{Experimental Setup}\label{sec:methods}
\subsection{Training Set}
We use in this work the Fisher Dataset \citep{Cieri2004TheFC}, a conversation corpus consisting of more than 16,000 English telephone conversations
averaging ten-minutes in duration and focusing on various topics.
The audio was recorded separately in two channels resulting in 2000 hours of transcribed speech.\footnote{The transcription was done using the Quick Transcription specification \citep{Cieri2004TheFC}, resulting in some inaccuracies and untranscribed portions. Here, we only used the transcriptions to obtain speech segments containing vocal activity to train the HifiGan and HuBERT model. The DLM was trained on the unsegmented raw data.}

For the training of HuBERT and HifiGAN models, we follow the preprocessing steps of \citet{nemo2019}\footnote{https://gitlab.nrp-nautilus.io/ar-noc/nemo/-/blob/master/scripts/process\_fisher\_data.py} to obtain a collection of single-channel voice segments of the Fisher dataset. The segments vary mostly from 10--15 seconds, with a total duration of about 1800 hours.
We divide the Fisher dataset into train/valid/test sets with a 98/1/1 split (different speakers in each split).

\subsection{Model Training}
We train a HuBERT Base model~\cite{weining2021hubert} from raw audio. The encoder contains seven 512-channel CNN layers with strides [5,2,2,2,2,2,2] and kernel widths [10,3,3,3,3,2,2], converting the signal rate from 16000 samples/sec down to 50 frames/sec. It is followed by 12 Transformer blocks. The model is trained with a masked objective for 3 iterations following the same recipe as in \cite{weining2021hubert}.
The model alternates between feature extraction/quantization and masked-prediction training in each iteration. We used the k-means algorithm with codebook sizes of 100, 500, and 500 to quantize the MFCC features, the 6th transformer layer features, and the 9th transformer layer features for the three HuBERT training iterations. After training, we quantize the final transformer layer features into 500 units for the DLM training. We choose a large codebook size of 500 to model various kinds of vocalizations beyond broad phonetic classes. Following \citet{weining2021hubert}, we use 250k training updates in the first iteration and 400k model updates in subsequent training iterations using 32 V100 32GB GPUs. As the transformer does not change the input frame rate, the encoded discrete units have a frame rate of 50 units per second (one every 20ms).
We show in Table \ref{SC-tab:FisherABX} that our HuBERT model trained on the Fisher dataset learns better phonetic information suitable for conversations than the publicly available HuBERT model trained on audiobooks \cite{weining2021hubert}.

We train the HifiGAN model on a small subset of the Fisher dataset segments consisting of 120 speakers with 10 minutes each.
These speakers were selected to be of high intelligibility using the average perplexity of a phone recognizer trained on the clean Librispeech 100h training subset \cite{riviere2021towards}.
The model is trained to generate the audio waveform given HuBERT units of a segment and a speaker embedding vector.

For the DLM models, we use a transformer model consisting of 6 layers, with 8 attention heads per layer, and an embedding size of 512. 
When cross-attention is used, it is added to the top 4 transformer layers.
We show in Table \ref{SC-tab:n_cross_layers} the effect of number of cross-attention layers on language modeling metrics.
We train the DLM model on the parallel unit streams encoded from 2000 hours of stereo audio, each sample contains up to 6144 unit pairs, an equivalent of 123 seconds. The models are trained on a total of 32 V100 32GB GPUs, with a batch size of 370 seconds of audio per GPU for a total number of 250k steps. We used an Adam optimizer \cite{kingma2014adam} with a max learning rate
of $5 \times 10 ^{-4}$. The implementation of the DLM model is done using the fairseq~\cite{ott2019fairseq} toolkit. It took us 66 hours on average to train 100k steps of DLM models without edge unit prediction, and 95 hours with additional edge unit prediction objective.

We also train a Multi-Stream Transformer Language Model (MS-TLM, \citealp{Kharitonov2021pgslm}), a single transformer model taking two streams of units as input and autoregressively predict the next units in both streams. It is a standard Transformer Language Model, with 6 layers, 8 attention heads per layer and an embedding size of 512, with the difference that the embedding layer concatenates the two embeddings of the two parallel units, and the output layer produces two softmax layers to predict the next units in both streams. We train the MS-TLM model similarly to the DLM models as previously mentioned. Training 100k steps of MS-TLM model took us 40 hours.

For the cascaded system, we use a pre-trained ASR model\footnote{\label{ASRmodel}We use the robust wav2vec2-large \href{https://github.com/facebookresearch/fairseq/tree/main/examples/wav2vec\#pre-trained-models}{model} fine-tuned on Switchboard dataset \cite{hsu21_interspeech}. For decoding, we use the 4-gram KenLM language model trained on Switchboard dataset.} to decode the Fisher dataset.
We then train a standard 6-layer Transformer Language Model on the turn-based conversations obtained from the ASR. We pre-process the text using a byte pair encoding (BPE, \citealp{sennrich-etal-2016-neural}) with 20k iterations and limit each sample to have 512 tokens. We trained the language model for 100k steps on 32 V100 32GB GPUs with a batch size of 2048 tokens per GPU. We use the same optimizer as for other models. Finally, we use the Google TTS API to synthesize generated conversations, with two different voices indicating two different speakers.


\subsection{Evaluation Metrics}\label{sec:evals}
This section presents the evaluation metrics used to assess our dialogue models on two dimensions: Training and Generation.

\subsubsection{Training Metrics}
These metrics evaluate the dialogue modeling performance in each channel separately using metrics close to the training loss. They are computed by encoding files from the development set and extracting statistics on the predicted outputs at each time steps. They are used to compare the different versions of the DLMs and therefore not applied to the cascaded model. 

\paragraph{Edge Unit Prediction.} We report the Negative Log-Likelihood (NLL), or Cross Entropy loss when predicting edge units. We also compute the Prediction Accuracy.

\paragraph{Edge Duration Prediction.} We use the Mean Absolute Error (MAE), or L1 Loss when evaluating edge duration prediction (a MAE of 1 corresponds to 20ms of error). The Duration Accuracy is also reported.


\begin{figure*}[ht]
     \centering
     \includegraphics[width=1\textwidth]{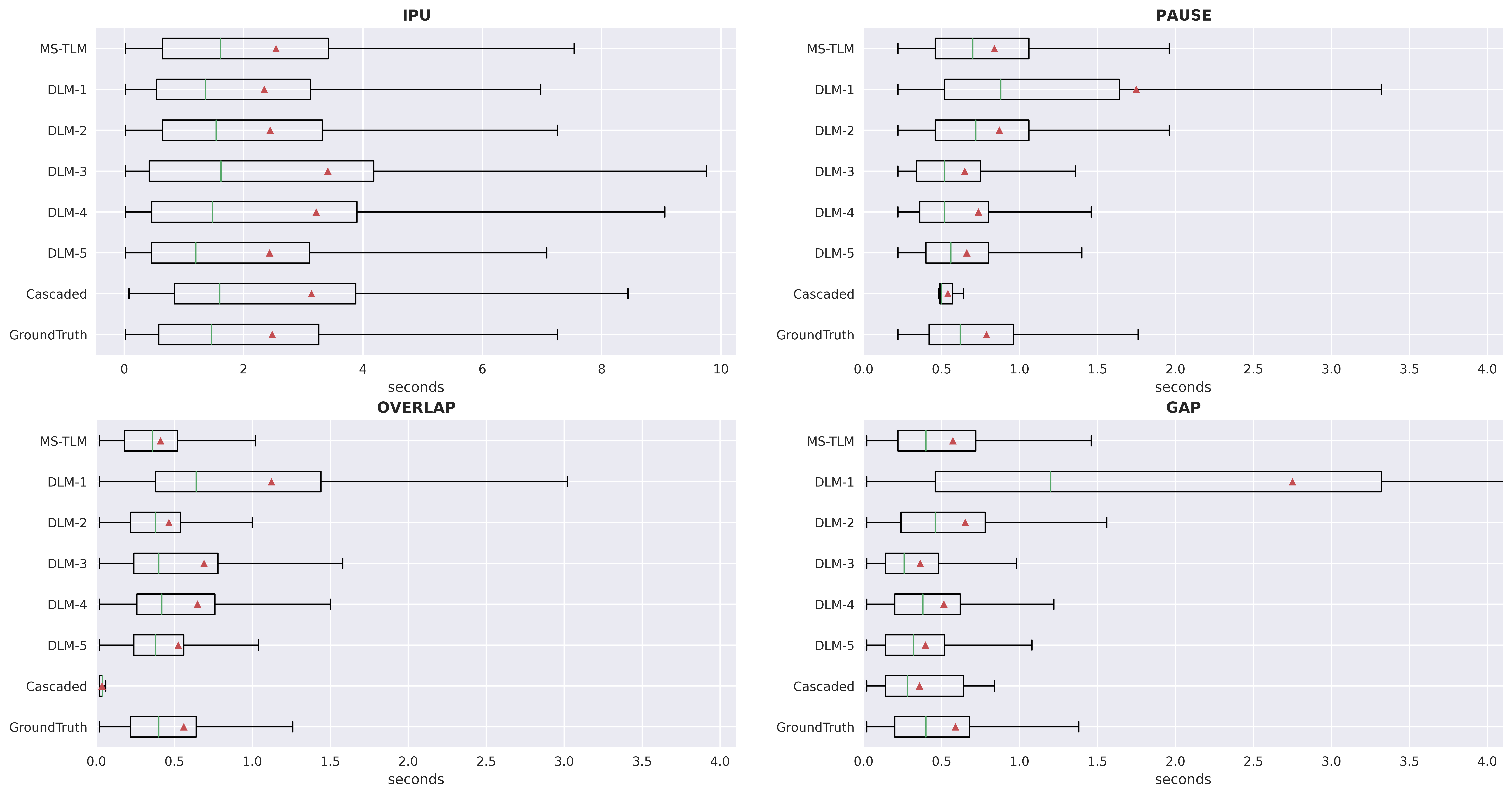}
     \caption{\textbf{Distributions of durations of turn-taking events} in prompted continuations across models, compared to the prompts' continuation ground truth segments (see models ids in Table \ref{tab:eval_modeling}). The green line and the red triangle represent the mean and the median of the events respectively.}
     \label{fig:states_statistics}
\end{figure*}

\subsubsection{Dialogue Generation Metrics}
We evaluate the generation properties of our models using descriptive statistics, automatic metrics and human-based judgements. 
Unless otherwise written, we perform conditional generation and generate 90-second long continuations using 117 30-second long prompts extracted from the development set and use the default generation temperature of 1.0. We generate the units by sampling among the top 20 possible units.

\paragraph{Turn-taking Event Statistics.} We compute the turn-taking events as defined in section \ref{sec:turn-taking-definition} using the samples generated by the models with a Voice Activity Detection (VAD) using pyannote library\footnote{https://github.com/pyannote/pyannote-audio} \cite{Bredin2020pyannote}. We then analyse the statistics of these turn-taking events (number of events and their durations) across different models.


\paragraph{Turn-taking Event Consistency.} We evaluate the model's capacity to generate consistent conversations in terms of turn-taking events. We measure the Pearson correlation between the total duration of events in each prompt and in the corresponding continuation.

\paragraph{Natural Dialogue Event Statistics.} 
We evaluate the naturalness of the generated speech 
by focusing on the Speaking Rate (WPM, words per minute), Laughter Frequency (LPM, laughs per minute), Filler Word Rate (FWR, filler words per 100 words) and Floor Transfer Offset (duration between two consecutive turns of the two speakers, a positive FTO represents a gap while a negative FTO represents an overlap). For this evaluation, we use the same ASR model used to decode the Fisher dataset\textsuperscript{\ref{ASRmodel}} to transcribe the generated speech. To detect laughs in the speech, we use an open-source model described in \citet{gillick21_interspeech}.\footnote{https://github.com/jrgillick/laughter-detection} To compute the FWR, we use the following filler words set: \{'uh', 'um', 'like', 'i mean', 'you know'\}.

\paragraph{Semantic Evaluation.} 
We use two evaluation metrics proposed in \citet{kushal2021gslm}, perplexity (PPL) and VERT, to assess the generation quality and diversity of the models.
We first transcribe the generated speech using the ASR system. As these metrics are calculated on text sequences, we combine the text from two channels into a single turn-based text sequence\textsuperscript{\ref{textturn}}, ignoring any turns that are completely contained inside an other turn. We employ the open-source DialoGPT model\footnote{https://huggingface.co/microsoft/DialoGPT-medium} \cite{DialoGPT} to compute the perplexity on the turn-based sequences. We simply replace the speaker tokens (\texttt{<A>}, \texttt{<B>}) with the \texttt{<|endoftext|>} token, indicating a turn switch.
For the VERT metrics, we also compute the self-BLEU and auto-BLEU on the turn-based text sequences. As the conversation texts contain a lot of repetitions, we report the VERT-4 score instead of VERT-2 score as in \citet{kushal2021gslm}.

\begin{figure}
    \centering
    \includegraphics[width=.5\textwidth]{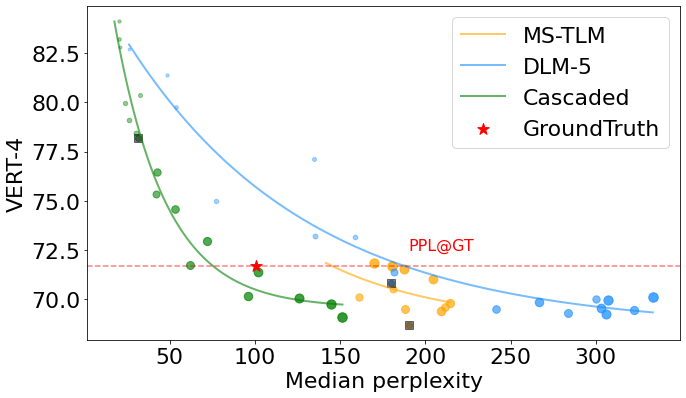}
    \caption{PPL vs VERT scores with unconditioned generation for MS-TLM, DLM-5 and Cascaded models compared to ground truth transcriptions. The sizes of the points correspond to the temperature used for generation (0.3--2.0), squares mean default temperature 1.0. The turn-based sequences are limited to 50 words.}
    \label{fig:pplvsvert}
\end{figure}

Since the PPL and VERT scores highly depend on the generation temperature, we perform generation on different temperatures ranging from 0.3--2.0. We then compute the PPL and VERT for each temperature and fit the points corresponding to different temperatures with an exponential line and report the PPL@GT (PPL with respect to the ground truth VERT) score
(cf. Figure \ref{fig:pplvsvert}). 
For the conditional generation case, we compute instead the conditional perplexity (cond. PPL), which is the perplexity of the generated sequence given the concatenation of the prompt sequence and generated sequence as input to the DialoGPT model.

\begin{table}[t]
\caption{\textbf{Training Metrics} across the DLM models that differ in Cross-Attention Layer (CA), Edge Unit Prediction (EP), Duration Prediction (DP) and Duration Delayed Factor ($\Delta$). The MS-TLM model used a single transformer with two input and output streams.
}
\label{tab:eval_modeling}
\centering
\resizebox{\columnwidth}{!}{
\begin{tabular}{c cccc | c@{\hspace{0.5\tabcolsep}}c c@{\hspace{0.5\tabcolsep}}c}
 \toprule
 & & & & & \multicolumn{2}{c}{\textbf{Edge Unit}} & \multicolumn{2}{c}{\textbf{Duration}} \\
Id & CA & EP & DP & $\Delta$ & NLL$\downarrow$ & Acc$\uparrow$ & MAE$\downarrow$ & Acc$\uparrow$ \\
\midrule
\midrule
& \multicolumn{4}{l}{\textbf{MS-TLM}}\\
0 & -  & - & - & - & 3.05 & 34.14 & - & - \\
& \multicolumn{4}{l}{\textbf{DLM}}\\
1 & \myXmark & \myXmark & \myXmark & - & 3.07 & 34.13 & - & - \\
2 & \cellcolor{mycell}\myVmark & \myXmark & \myXmark & - & 2.95 & 35.68 & - & - \\
3 & \myVmark & \cellcolor{mycell}\myVmark & \myXmark & - & 2.49 & 48.36 & - & - \\
4 & \myVmark & \myVmark & \cellcolor{mycell}\myVmark & 0 & 2.26 & 54.09 & 1.47 & 51.90 \\
5 & \myVmark & \myVmark & \myVmark & \cellcolor{mycell}1 & \textbf{2.25} & \textbf{54.27} & \textbf{1.23} & \textbf{58.18}  \\
 \bottomrule
\end{tabular}
}
\end{table}

\begin{table*}[t]
\caption{\textbf{Number of turn-taking events and cumulated durations per minute} across models for prompted continuations, compared to ground truth continuations, and to the same statistics in the training set.}
\label{tab:states_per_minutes}
\centering
\resizebox{1.7\columnwidth}{!}{
\begin{tabular}{c l | cccc | cccc}
\toprule
 & & \multicolumn{4}{c|}{\textbf{Number of occurrences / min}} & \multicolumn{4}{c}{\textbf{Cumulated duration /min}} \\
Id & Model & IPU & Pause & Gap & Overlap & IPU & Pause & Gap & Overlap \\
\midrule
\midrule
0 & \textbf{MS-TLM} & 19.4 & 10.6 & 5.1 & 3.3 & 49.4s & 8.9s & 2.9s & 1.3s \\
\midrule
1 & \textbf{DLM-1} & 17.7 & 7.9 & 3.9 & 5.5 & 41.4s & 13.8s & 10.7s & 6.1s \\
2 & \textbf{DLM-2} & 20.0 & 10.4 & 5.5 & 3.6 & 48.9s & 9.1s & 3.6s & 1.7s \\
3 & \textbf{DLM-3} & 19.0 & 1.8 & 4.9 & 11.7 & 65.0s & 1.1s & 1.8s & 8.1s \\
4 & \textbf{DLM-4} & 18.9 & 3.2 & 5.6 & 9.4 & 60.7s & 2.4s & 2.9s & 6.1s \\
5 & \textbf{DLM-5} & 24.2 & 5.4 & 7.2 & 10.9 & 59.1s & 3.6s & 2.9s & 5.8s \\
\midrule
6 & \textbf{Cascaded} & 17.5 & 0.0 & 14.9 & 0.0 & 54.8s & 0.0s & 5.3s & 0.0s \\
\midrule
  & \textbf{Ground Truth} & 21.6 & 7.0 & 7.5 & 6.5 & 53.5s & 5.5s & 4.4s & 3.6s \\
  & \textbf{Training Set} & 25.9 & 7.2 & 8.6 & 10.0 & 54.5s & 5.6s & 4.6s & 4.7s \\
\bottomrule
\end{tabular}
}
\end{table*}

\paragraph{Human Opinion Score.}
We perform a human evaluation on the generated examples. The opinions are based on two dimensions: \textit{N-MOS (naturalness Mean Opinion Score)} representing naturalness and turn-taking conversationality, and \textit{M-MOS (meaningfulness Mean Opinion Score)} for meaningfulness and content quality. For N-MOS, we asked the participants to concentrate on the
fluidity and naturality of the interaction as well as the expressiveness of the speakers regardless of meaning. For M-MOS, they should focus on what is being said and if it is semantically coherent. For these two evaluations, we used a scale of 1-5 (1: worst, 5: best).
The CrowdMOS package \cite{ribeiro2011crowdmos} was used for all subjective evaluations using the recommended recipes
for detecting and discarding inaccurate scores.
Indeed, we remove all workers whose correlation with the mean scores is lower than 0.25, and then filter out outlier workers whose correlation with the mean scores is lower than 0.6.
We enforced at least six raters for each of the generated samples. Participants were recruited using a crowd-sourcing platform.


\section{Results}

\subsection{Content and Duration Modeling}
Table \ref{tab:eval_modeling} reports the modeling evaluation metrics on our development subset of the Fisher dataset. In rows Id 1-5, we compare different DLM models, while row Id 0 represents the MS-TLM model, which takes as input multiple unit streams from different channels, and predicts the next-step units only. We note that for models Id 1-3, the next-step unit prediction objective is also included in the training process, but when the duration prediction objective is employed (models Id 4-5), the next-step unit prediction objective is omitted.

We observe that by using the self cross-attention layers, the edge unit prediction metrics slightly improve ($u$ NLL: 3.07 vs 2.95). On considering models Id 2 \& 3, we observe a huge improvement in edge unit NLL \& Accuracy when introducing the edge unit prediction objective ($u$ NLL: 2.95 vs 2.49). By introducing the duration prediction objective and removing the next-step unit prediction objective, we see that the model performs even better on the edge unit prediction metrics ($u$ NLL: 2.26), and finally the duration metrics greatly improves when we apply a delayed duration prediction ($d$ MAE: 1.47 vs 1.23).

On comparing with the MS-TLM model, we see that our best DLM model perform much better on content modeling. The reason, we believe, is related to the entangled modeling of content and duration in the MS-TLM model. 

\begin{figure*}[ht]
     \centering
     \includegraphics[width=0.8\textwidth]{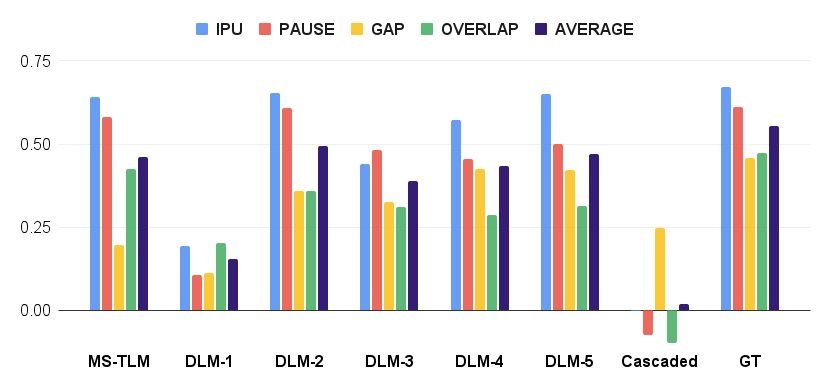}
     \caption{\textbf{Correlation between the duration of events in the prompts and in the continuations} across models, compared to ground truth (GT), where the correlation is computed between the first 30 seconds and the following 90 seconds of the samples.}
     \label{fig:conversation_consistency}
\end{figure*}

\begin{table}[t]
\caption{\textbf{Natural Dialogue Event Statistics.} Speaking Rate (WPM, words per minute), Laughter Frequency (LPM, laughs per minute) and Filler Word Rate (FWR, filler words per 100 words) of the prompted continuation speech across models, compared to ground truth continuations.}
\label{tab:natural_statistics}
\centering
\resizebox{0.9\columnwidth}{!}{
\begin{tabular}{c l | ccc}
\toprule
Id & Model & WPM & LPM & FWR\\
\midrule
\midrule
0 & \textbf{MS-TLM} & 139.17 & 1.88 & 9.36 \\
\midrule
1 & \textbf{DLM-1} & 123.60 & 1.98 & 9.39 \\
2 & \textbf{DLM-2} & 141.09 & 2.06 & 10.36 \\
3 & \textbf{DLM-3} & 281.41 & 7.08 & 3.40 \\
4 & \textbf{DLM-4} & 244.13 & 6.05 & 3.38 \\
5 & \textbf{DLM-5} & 211.98 & 3.62 & 5.50 \\
\midrule
6 & \textbf{Cascaded} & 216.73 & 0.00 & 7.08 \\
\midrule
  & \textbf{Ground Truth} & 181.46 & 3.60 & 7.25 \\
\bottomrule
\end{tabular}
}
\end{table}

\subsection{Turn-taking Event Statistics}
In this section, we analyse the distribution of the turn-taking events (as described in section \ref{sec:turn-taking-definition}) in the dialogue continuations generated by our models.
The statistics are computed over 3 hours of generated speech per model.

Figure \ref{fig:states_statistics} shows the distribution of each of the 4 turn-taking events: IPU, pause, gap and overlap. In this figure, the Ground truth corresponds to the true continuation of the prompts in the original corpus. Despite having a reasonably good modeling score (cf. Table \ref{tab:eval_modeling}), DLM-1, which has no cross-attention layers between the two transformer towers, has poor performance on turn-takings events, except for the IPU event. The lack of communication between the two channels during generation creates huge gaps and overlaps in the generated samples. 
The MS-TLM and DLM-2 models have similar distributions of shorter overlaps and longer pauses and gaps. They were trained using the next-step prediction loss on duplicated unit sequences, which could lead to repeated unit generation, causing a slow pace and more extended silences in the generated audio. The opposite effect happens when we introduce the edge unit prediction (DLM-3-5). These models manage to generate more overlaps, with pauses and gaps of shorter duration. These observations are further reinforced in Table \ref{tab:states_per_minutes}, which details the number of events and their total durations per minute. It is interesting to note that all models, except DLM-1, manage to capture the empirical fact that \textit{intra-turn} pauses tend to be longer than \textit{between-turn} gaps \cite{brady1968statistical,ten2005temporal,heldner2010pauses}. 

The cascaded model only produces alternating speech turns and therefore has almost no overlap and pause. This also results in low variance in the gap distribution, making the geneation sounds like machine conversation.



\subsection{Turn-taking Event Consistency}

Figure \ref{fig:conversation_consistency} shows the correlation between the total duration of turn-taking events in the prompts and in the generated continuations. For the ground truth, we compute the correlation of the events' duration between the first 30 seconds and the folowing 90 seconds in each sample. We observe that in general all models except DLM-1 and cascaded have good correlations, showing their ability to maintain the dialogue consistency. Unsurprisingly, the cascaded model has no correlation with the prompt events, except for the gaps, which are proportional to the number of turn changes.

\begin{figure}
    \centering
    \includegraphics[width=.4\textwidth]{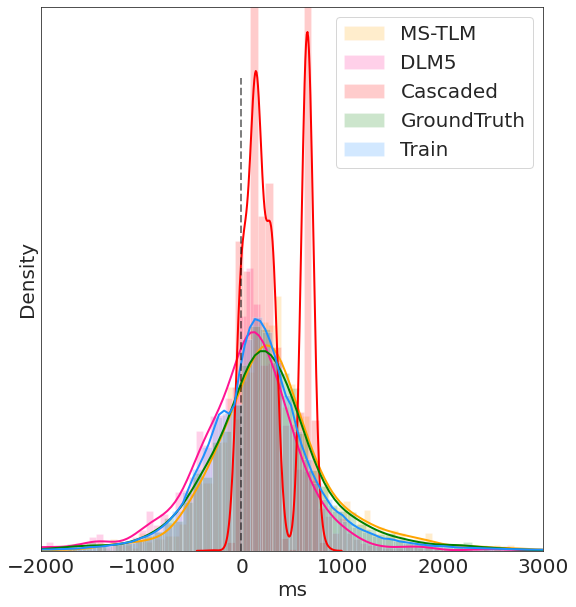}
    \caption{Histogram of Floor Transfer Offset (FTO) in the generated speech across models, compared to ground truth continuations and the training set.}
    \label{fig:floor_transfer_offset}
\end{figure}

\subsection{Natural Dialogue Event Statistics}
Table \ref{tab:natural_statistics} reports the naturalness statistics on the generated samples of our models. 
We first notice that, compared to ground truth, models that don't have edge unit prediction (MS-TLM, DLM-1--2) tend to produce speech with 
less information and more hesitations (lower rate, less laughter, more filler words)
than those with edge unit prediction (DLM-3--5). 
Adding duration prediction can effectively help to produce more natural speech, but it still produces more words than ground truth. 
The cascaded model is unable to produce laughter as the ASR and TTS modules are not able to capture these information, it also generate nearly "non-stop" speech at a faster rate than natural speech.
Looking at Figure \ref{fig:floor_transfer_offset}, we see indeed that the cascaded model has no negative FTO (overlap), and the positive FTOs (gaps) fall mostly in the range of one second. In general, other models seem to have good FTO distribution compared to the reference ground truth and training set.

\subsection{Semantic Evaluation}
For semantic metrics, we perform both conditional and unconditional generations. For conditional generation, we select 50 10-second long prompts in the validation set. For each model and temperature, we generate 50 samples and limit the transcribed turn-based text sequences to 50 words.

We found that certain models is not possible to obtain PLL@GT as they tend to generate repeated units at low temperatures, creating complete noise in the synthesis. We therefore report the PPL scores for the default temperature 1.0 (@t1). As shown in Table \ref{tab:automatic_eval}, we see that the dialogue models fail to generate semantically coherent speech, resulting in high perplexity, especially in prompted generation. 
The cascaded model has a very good perplexity as the language model was trained on word and sub-word levels, it even has a higher PPL@GT than the ground truth in the unconditional case. When it comes to conditional generation, the cascaded model has a good PPL, but is still way below the ground truth.

\begin{table}[t]
\caption{\textbf{Semantic Evaluation.} Perplexity of ASR-transcribed generated speech
at default temperature (@t1) and at ground truth VERT (@GT) in both unconditional and conditional generation across models compared to ground truth transcriptions. We limit the transcribed turn-based sequences to 50 words.}
\label{tab:automatic_eval}
\centering
\resizebox{1.\columnwidth}{!}{
\begin{tabular}{c l | cc | cc}
& & \multicolumn{2}{c|}{\textbf{unconditional}} & \multicolumn{2}{c}{\textbf{conditional}} \\
\midrule
& & \multicolumn{2}{c|}{PPL$\downarrow$} & \multicolumn{2}{c}{cond. PPL$\downarrow$} \\
Id & Model & @t1 & @GT & @t1 & @GT \\
\midrule
\midrule
0 & \textbf{MS-TLM} & 190.59 & 144.82 & 741.86 & - \\
\midrule
1 & \textbf{DLM-1} & 145.85 & - & 195.89 & - \\
2 & \textbf{DLM-2} & 218.30 & - & 453.73 & - \\
3 & \textbf{DLM-3} & 155.17 & 161.58 & 463.27 & 329.74 \\
4 & \textbf{DLM-4} & 290.07 & 231.00 & 693.48 & 314.49 \\
5 & \textbf{DLM-5} & 179.65 & 187.16 & 605.84 & 365.08 \\
\midrule
6 & \textbf{Cascaded} & 32.23 & 80.80 & 45.93 & 117.06 \\
\midrule
  & \textbf{Ground Truth} & 100.85 & 100.85 &  65.00 & 65.00 \\
\bottomrule
\end{tabular}
}
\end{table}

\subsection{Human evaluations. }
For this evaluation, we filter the prompts to contain genuine alternations between the two interlocutors and balanced gender. We retained 50 10-second long prompts and generated 10 20-second long continuations for each prompt. Human evaluation results are reported in Table \ref{tab:eval_mos}.
The naturalness and meaningfulness MOS scores correlate well with results in previous sections. 
The DLM-5 model has the best performance among dialogue models, while the DLM-1 performs significantly worse on both scores. Interestingly, whereas there is a large gap between our best model and ground truth on meaningfulness (1.73 points on the 5 points scale) this gap is much reduced on turn-taking (.53 points). 
The cascaded model shows a lack of naturalness, while having better scores on meaningfulness than all dialogue models. 
However, it is still far below the ground truth despite having a very good semantic scores.
Overall, our models can generate dialogues mimicking natural turn-taking, while fail maintaining cross-sentence meaningfulness. We believe the lack of semantic coherence in generated dialogues results from the fine-grained acoustic units used for modeling and the small training corpus size.


\begin{table}[t]
\caption{\textbf{Human Evaluations}. Conversation Naturalness (N-MOS) and Conversation Meaningfulness (M-MOS) on a 5 point scale (5 is best) with 95\% CI.}
\label{tab:eval_mos}
\centering
\resizebox{0.9\columnwidth}{!}{
\begin{tabular}{c l  cc}
\toprule
Id & Model &  N-MOS$\uparrow$ & M-MOS$\uparrow$ \\
\midrule
\midrule
0 & \textbf{MS-TLM} & 3.31 ± 0.43 &  2.29 ± 0.49 \\
\midrule
1 & \textbf{DLM-1} & 2.25 ± 0.60 &  1.70 ± 0.44 \\
2 & \textbf{DLM-2} & 2.95 ± 0.37 &  2.24 ± 0.47 \\
3 & \textbf{DLM-3} & 3.29 ± 0.43 &  2.20 ± 0.44 \\
4 & \textbf{DLM-4} & 3.36 ± 0.44 &  2.18 ± 0.46 \\
5 & \textbf{DLM-5} & 3.70 ± 0.46 &  2.48 ± 0.49 \\
\midrule
6 & \textbf{Cascaded} & 2.38 ± 0.63 &  2.70 ± 0.38\\
\midrule
& \textbf{Ground Truth} & 4.23 ± 0.26 &  4.21 ± 0.25 \\
\bottomrule
\end{tabular}
}
\end{table}


\section{Conclusion and Future Work}
We have presented dGSLM, the first model for spoken dialogue generation trained from raw audio. This model has been shown to reproduce naturalistic intelligible speech, while trained on only 2k hours of audio from telephone conversations. Informal inspection of the generated samples
\textsuperscript{\ref{demowebsite}}
shows that it is able to reproduce non-verbal vocalizations (laughter, backchannels). Detailed analysis of the turn-taking events show that the model is able to reproduce accurate synchronization including distribution and duration of turn-taking events like IPU, gaps, pauses and overlaps. In particular, it is able to reproduce the rather puzzling observation that inter-turn pauses tend to be on average longer than between turn gaps, suggesting the pauses alone are not a sufficient signal to indicate a change of turn. 

Although the model lacks the ability to produce semantically coherent speech, it paves the way for the construction of more naturalistic human-machine dialogue systems. The logic and timing of turn-taking which has been up to now very difficult to model artificially emerges naturally from our system, while it is clearly not yet able to process speech at a deep semantic level. This indicates that a model that correctly predicts synchronization between turns can be learned from relatively a small amount of data. This is surprising given that one major paralinguistic information, intonation, was not explicitely encoded in the input (or the output) of the system. Further work incorporating pitch \cite{Kharitonov2021pgslm} could potentially improve the current results. 
Results from the cascaded system also suggest that 
either using larger linguistic units (like BPE) from raw audio \cite{borsos22_audiolm} or combining our model with text-based models would create systems which could generate more natural and meaningful conversations.


\newpage
\appendix

\begin{center}
\textbf{\large Appendix}
\end{center}
\setcounter{equation}{0}
\setcounter{figure}{0}
\setcounter{table}{0}
\makeatletter
\renewcommand{\theequation}{A\arabic{equation}}
\renewcommand{\thefigure}{A\arabic{figure}}
\renewcommand{\thetable}{A\arabic{table}}
\renewcommand{\bibnumfmt}[1]{[A#1]}
\renewcommand{\citenumfont}[1]{A#1}
\section{Phonetic Quality of HuBERT Fisher}\label{sec:appendix_hubert}
In Table \ref{SC-tab:FisherABX}, we compare the HuBERT Base model  \cite{weining2021hubert} trained on 2000h of Fisher dataset versus 1000h of Librispeech dataset on the machine-ABX phonetic test. We used Libri-light ABX \cite{kahn2020librilight} for the Lirispeech test. For the Fisher, we generated a Fisher ABX dataset using the phonetic alignments obtained from Fisher development set. The results clearly show a domain effect, whereby the Fisher dataset is a better training set than the Librispeech dataset for ABX discriminations in Fisher. 
\begin{table}[ht]
\caption{\textbf{Within and Across-Speaker ABX error} on Fisher dev and LibriSpeech dev-clean datasets for HuBERT Base and HuBERT Fisher models.}
\label{SC-tab:FisherABX}
\centering
\resizebox{\columnwidth}{!}{
\begin{tabular}{l|cc | cc}
\toprule
& \multicolumn{2}{c|}{Fisher} & \multicolumn{2}{c}{LibriSpeech} \\
&  within$\downarrow$ & across$\downarrow$ &  within$\downarrow$ & across$\downarrow$  \\
\midrule
\midrule
HuBERT Base & 7.77 & 12.57 & \textbf{3.95} & \textbf{4.69}  \\
HuBERT Fisher & \textbf{5.50} & \textbf{8.35} & 11.17 & 14.70 \\
\bottomrule
\end{tabular}
}
\end{table}

\section{Effects of Cross-Attention Layer}

In Table \ref{SC-tab:n_cross_layers}, we show the NLL and Accuracy scores of Transformer Language Models as a function of number of Cross-Attention layers. 
The models are Two-tower Transformer Systems but are trained with only Next-Step Prediction Objective. 
We find that more layers give better scores, but that 4 layers of cross-attention gives almost the same performance as 6 for less complexity. 
\begin{table}[ht]
\caption{\textbf{Unit Prediction loss (NLL) \& Accuracy metrics} of DLM models trained with different number of cross-attention layers. When the number of cross-attention layers is less than 6, they are put on top of self-attention layers. The models are trained with the Next-step Unit Prediction Objective on the parallel unit streams of the Fisher stereo audio dataset.}
\label{SC-tab:n_cross_layers}
\centering
\resizebox{0.6\columnwidth}{!}{
\begin{tabular}{c|cc}
\toprule
n cross layers&  NLL$\downarrow$ & Acc$\uparrow$ \\
\midrule
\midrule
0/6 & 1.387 & 71.77 \\
2/6 & 1.341 & 72.06 \\
4/6 & \textbf{1.338} & \textbf{72.10} \\
6/6 & \textbf{1.337} & \textbf{72.11} \\
\bottomrule

\end{tabular}
}
\end{table}
\clearpage
\newpage
\bibliography{main_dGSLM}
\bibliographystyle{acl_natbib}

\end{document}